\documentclass[pdflatex,sn-mathphys-num]{sn-jnl}
\usepackage{graphicx}%
\usepackage{multirow}%
\usepackage{amsmath,amssymb,amsfonts}%
\usepackage{amsthm}%
\usepackage{mathrsfs}%
\usepackage[title]{appendix}%
\usepackage{xcolor}%
\usepackage{textcomp}%
\usepackage{manyfoot}%
\usepackage{booktabs}%
\usepackage{algorithm}%
\usepackage{algorithmicx}%
\usepackage{algpseudocode}%
\usepackage{listings}%

\begin{document}

\title[Article Title]{Investigation of D-Wave quantum annealing for training Restricted Boltzmann Machines and mitigating catastrophic forgetting}

\author[1]{\fnm{Abdelmoula} \sur{El-Yazizi}}\email{ae897@msstate.edu}
\equalcont{These authors contributed equally to this work.}

\author*[1]{\fnm{Yaroslav} \sur{Koshka}}\email{ykoshka@ece.msstate.edu}
\equalcont{These authors contributed equally to this work.}

\affil[1]{\orgdiv{Electrical and Computer Engineering}, \orgname{Mississippi State University}, \orgaddress{\city{Mississippi State}, \postcode{39762}, \state{MS}, \country{USA}}}

\abstract{Modest statistical differences between the sampling performances of the D-Wave quantum annealer (QA) and the classical Markov Chain Monte Carlo (MCMC), when applied to Restricted Boltzmann Machines (RBMs), are explored to explain, and possibly address, the absence of significant and consistent improvements in RBM trainability when the D-Wave sampling was used in previous investigations. A novel hybrid sampling approach, combining the classical and the QA contributions, is investigated as a promising way to benefit from the modest differences between the two sampling methods. No improvements in the RBM training are achieved in this work, thereby suggesting that the differences between the QA-based and MCMC sampling, mainly found in the medium-to-low probability regions of the distribution, which are less important for the quality of the sample, are insufficient to benefit the training. Difficulties in achieving sufficiently high quality of embedding RBMs into the lattice of the newer generation of D-Wave hardware could be further complicating the task. On the other hand, the ability to generate samples of sufficient variety from lower-probability parts of the distribution has a potential to benefit other machine learning applications, such as the mitigation of catastrophic forgetting (CF) during incremental learning. The feasibility of using QA-generated patterns of desirable classes for CF mitigation by the generative replay is demonstrated in this work for the first time. While the efficiency of the CF mitigation using the D-Wave QA was comparable to that of the classical mitigation, both the speed of generating a large number of distinct desirable patterns and the potential for further improvement make this approach promising for a variety of challenging machine learning applications. }


\maketitle

\section{Introduction} \label{sec1}
In recent years, advancements in quantum computing (QC) hardware have paved the way for the emerging field of quantum machine learning (QML), currently implemented on noisy intermediate-scale quantum (NISQ) computers. While gated quantum computers attract most of the interest due to their potential to eventually enable a general-purpose QC, it is a Quantum Annealer (QA), specifically the QA from D-Wave, Inc., that became the world’s first commercial QC. This head start is at least in part responsible for the significant attention the QAs have received for various applications, including QML.  In one of the early attempts, relevant to this study, the D-Wave QA showed promise of efficient sampling to train undirected probabilistic graphical models, specifically deep Boltzmann Machines (BMs) \cite{adachi_application_2015}. The reasons behind that early improvement reported in Ref. \cite{adachi_application_2015} have not been conclusively established. Afterwards, many studies attempted to use the D-Wave for sampling in training Restricted Boltzmann Machines (RBMs) but failed to achieve substantial improvement \cite{benedetti_estimation_2016,korenkevych_benchmarking_2016,rocutto_quantum_2021,dixit_training_2021}. 

The improvements in the QA hardware during the past decade have resulted in an increase in the number of qubits and, with the introduction of the Pegasus architecture, in an increase in the connectivity between qubits. This has enabled embedding larger graphs. However, with the improvement of the hardware, the task of using the QA for RBM training faces several questions, both new and longstanding, including the following: (1) whether the new QA architecture (Pegasus) enables the quality of embedding that is at least comparable to the older D-Wave versions, (2) whether the D-Wave QA can offer sampling performance that is at least comparable to classical methods, and (3) in the case that the sampling performance is different from the classical (better or worse), what reasons are behind those differences. The much larger number of qubits alone allows achieving higher connectivity by combining multiple qubits into a single logical unit; however, this chaining may cause deviation from the intended state of the logical units. Furthermore, as was shown in previous studies \cite{pelofske_comparing_2023} and discussed in the previous work by the authors’ group \cite{elyazizi1_compare}, the ability of the D-Wave QA to do its primary task, which is finding the Ground State (GS), continues to improve. As was shown in Ref. \cite{pelofske_comparing_2023}, the newer generations of the QA show the trend of finding the GS more frequently. Naturally, as the frequency of finding the GS as a result of a certain number of annealing repetitions increases, the diversity of other sampled solutions diminishes. This reduction in sample variety could negatively impact the QA’s effectiveness as a sampler. 

Many previous studies reported attempts to compare D-Wave and Markov Chain Monte Carlo (MCMC) samples using rigorous statistical methods, such as the Kullback-Leibler (KL) divergence \cite{goto_online_2023}, or the expectation values of selected observables whose operators are diagonal in the computational basis \cite{gonzalez_testing_2021}. In contrast, earlier work by the authors’ group focused on understanding the differences between D-Wave-based sampling and classical MCMC sampling, using a particular criterion that was found by the authors to be useful for analyzing sampling from the probability distribution of RBMs \cite{koshka_toward_2020,koshka_comparison_2020,koshka_comparison_2021}. The comparison was centered on local valleys (LVs) in the RBM energy function found by each of the two sampling techniques. In the authors’ most recent work \cite{elyazizi1_compare}, this latter comparison was performed specifically under sampling conditions similar to those used in classical Contrastive Divergence (CD-\textit{k}) of the RBM training (specifically, CD-\textit{1}, with $k = 1$ and $T = 1$), where \textit{T} denotes the temperature in the Boltzmann distribution and \textit{k} is the number of steps in the Gibbs chain. For probabilistic graphical models, including RBM, many LVs are formed in the configuration space during training. Some of those LVs have local minima (LMs) that reside near one or another training pattern (TP). A sampler that repeatedly misses many of these LVs could cause poor learning of the class corresponding to those missed LVs. Moreover, RBM training is known to create “spurious” LVs containing undesirable low-energy (i.e., high-probability) states that must be properly sampled during training to either remove the corresponding LV or assign low probability to those states. However, the high probability of those LMs does not guarantee that the states inside the corresponding LV will be consistently included in the sample according to their high probability. For example, a deep LV may have a narrow basin of attraction \cite{koshka_comparison_2020}, which reduces the likelihood that the sampler will access states within that LV. 

Statistical analysis centered on LVs has revealed useful differences between the D-Wave and classical sampling methods \cite{elyazizi1_compare}. Specifically, the D-Wave and the Gibbs sampling techniques were compared when sampling from a classically trained RBM. The samples were compared by the number of LVs to which they belonged and the energy of the corresponding LMs. Many of the LVs found by the two techniques differed. Many potentially important LVs were found by only one sampling technique, while missed by the other. The results of this previous work communicated a reasonable optimism that the D-Wave and the classical Gibbs sampling could complement each other when used in RBM training, particularly by contributing statistically relevant information from different regions of the configuration space, at least in the medium- to high-energy ranges (which means medium- to low-probability of observing that state). However, the substantial overlap observed in the previous work between the D-Wave-based and Gibbs sampling in the low-energy (high-probability) regions may imply that D-Wave does not provide a substantial difference, possibly explaining the limited improvements observed in many previous attempts to enhance RBM trainability using the QA \cite{benedetti_estimation_2016,korenkevych_benchmarking_2016,rocutto_quantum_2021,dixit_training_2021}.

In this work, we continued investigating the potential of using the D-Wave QA as a generative model, specifically for two ML applications based on RBMs: (1) the RBM training and (2) the mitigation of Catastrophic Forgetting (CF) during Incremental Learning (IL). The common feature applied to both applications was the use of the D-Wave’s QA to generate samples from the RBM’s probability distribution. 

There were three main goals in this study. The first goal was to investigate the implications of transitioning to the new Pegasus lattice of the D-Wave QA. This lattice enables the embedding of substantially larger graphs, approaching, if not yet reaching, the scales needed for practical RBM applications. However, while the number of qubits is significantly larger (and continues to grow) in Pegasus compared to the previous versions, the increase in the connectivity between qubits is more modest and is expected to scale slower than the number of qubits. This means that the increase in the number of the logical units in problems to be solved on the D-Wave must rely primarily on chaining more qubits than in the previous studies that were using earlier versions of the hardware (chaining means combining multiple qubits to represent one logical unit). This may lead to a degradation in the quality of embedding, especially for RBM. This motivated the first goal of this work: to establish whether the QA can provide RBM training performance that is at least comparable to classical methods (if not an improvement) when the Pegasus hardware is used. 

The second goal was to verify a hypothesis that consisted of two parts. In the first part, it is suggested that the modest or no improvements in the previous attempts to use the D-Wave in RBM training can be attributed to the presence of only modest differences, but also significant statistical overlaps, between the classical MCMC and the D-Wave-based sampling methods. In the second part of the hypothesis, it is speculated that by combining the two techniques (i.e., the use of a hybrid D-Wave/classical sampling), it may be possible to explore the modest complementarity of the two sampling methods (i.e., the complementarity revealed in our previous work \cite{elyazizi1_compare}) and potentially achieve higher improvement from the D-Wave-based training. Alternatively, a lack of improvements from this approach would not necessarily serve as the final evidence of the limited potential of the D-Wave for this application. However, it would serve as another evidence supporting the hypothesis about why the improvements in the RBM training previously targeted by other groups were limited or absent. It should be noted that even in the case of such a negative outcome for this second goal of the work, the possibility of a comparable to classical sampling (i.e., training) efficiency would still be appealing, at least due to the sampling speed, i.e., the nearly instantaneous acquisition of a large sample when using a QA. 

The third goal of the study was to evaluate the feasibility of benefiting from the particular statistical differences between the QA-generated and classical samples by applying QA sampling to a different ML application involving RBMs. As discussed above, the differences between the D-Wave and the classical sampling were found predominantly for the regions of the configuration space having low or moderate probability states with low density of states. This property could benefit ML tasks where it is critical to generate important states, some of which may have an average-to-low overall probability of being sampled.  This motivated the third goal of this work, which was to use the D-Wave as a generative model for mitigating CF by the generative replay during IL of an RBM. 

The remainder of the paper is structured as follows. In Section \ref{sec:2}, the background information relevant to this work is presented, including challenges faced during RBM training, relevant details on MCMC methods, and adiabatic QA. In Section \ref{sec:3}, the methodology used in this work is described, including RBM embedding into the D-Wave lattice, RBM training using different methods, and the mitigation of CF during RBM IL. The results and discussion of the RBM training and CF mitigation are presented in Section \ref{sec:4}. Lastly, the paper is concluded by summarizing the key findings and suggesting future research opportunities.

\section{Background}\label{sec:2}
\subsection{Challenges in Training Restricted Boltzmann Machines}\label{sec:2.1}
RBM is a bipartite undirected probabilistic graphical model with a visible layer, which represents the training data, and a hidden layer, which captures the features in the data. During the learning of the RBM, the goal of the training algorithm is to adjust the weights and biases so that the model’s probability distribution becomes as close as possible to the data distribution (data to be learnt). In other words, the aim is to find the parameters ($\theta$) that maximize log-likelihood given the training data $\boldsymbol{\upsilon_{tr}}$: $\ln \mathcal{L}(\theta | \upsilon = \boldsymbol{\upsilon_{tr}}) = \ln p(\theta | \upsilon = \boldsymbol{\upsilon_{tr}})$. The joint probability distribution of the visible and hidden units in an RBM is: 
\begin{equation}
  p(\boldsymbol{\upsilon},\boldsymbol{h}) = \frac{1}{Z} e^{-\frac{E(\boldsymbol{\upsilon},\boldsymbol{h})}{T}}
\label{eq1}
\end{equation}

\noindent and the marginal probability distribution over the visible units is:
\begin{equation}
  p(\boldsymbol{\upsilon}) = \sum_{\boldsymbol{h}} p(\boldsymbol{\upsilon},\boldsymbol{h}) = \frac{1}{Z} \sum_{h} e^{-\frac{E(\boldsymbol{\upsilon},\boldsymbol{h})}{T}}
\label{eq2}
\end{equation}
were $Z$ is the partition function, $T$ is the temperature, and $E(\boldsymbol{\upsilon},\boldsymbol{h})$ is the energy of the RBM for the configuration of visible units $\boldsymbol{\upsilon}$ and hidden units $\boldsymbol{h}$.

The maximization of $\ln \mathcal{L}(\theta | \boldsymbol{\upsilon})$ is achieved by the gradient-descent-based optimization of the RBM parameters $\boldsymbol{\omega_{ij}}$,$\boldsymbol{b_j}$,$\boldsymbol{c_i}$. The gradients of the log-likelihood with respect to these parameters, given a single training pattern $\boldsymbol{\upsilon_{tr}}$ are: 
\begin{equation}
\frac{\partial \ln \mathcal{L}(\theta | \boldsymbol{\upsilon_{tr}})}{\partial \omega_{ij}} = p(H_i = 1 | \boldsymbol{\upsilon_{tr}}) \upsilon_j - \sum_{\upsilon} p(\boldsymbol{\upsilon}) p(H_i = 1 | \boldsymbol{\upsilon}) \upsilon_j
    \label{eq3}
\end{equation}

\begin{equation}
\frac{\partial \ln \mathcal{L}(\theta |  \boldsymbol{\upsilon_{tr}})}{\partial b_j} = \upsilon_j - \sum_{\upsilon} p(\boldsymbol{\upsilon}) \upsilon_j
    \label{eq4}
\end{equation}

\begin{equation}
\frac{\partial \ln \mathcal{L}(\theta | \boldsymbol{\upsilon_{tr}})}{\partial c_i} = p(H_i = 1 |  \boldsymbol{\upsilon_{tr}}) - \sum_{\upsilon} p(\boldsymbol{\upsilon}) p(H_i = 1 | \boldsymbol{\upsilon})
 \label{eq5}
\end{equation}
In Eq.~(\ref{eq3}), the first term is the expectation value under the conditional distribution given $\boldsymbol{\upsilon_{tr}}$, while the second term is the expectation value under the model distribution. The calculation of this second term requires summation over all values of visible units, which results in a significant computational complexity. The same computational burden applies to the second terms of Eqs.~(\ref{eq4}) and (\ref{eq5}). To alleviate this issue, the expectation terms under the model distribution are typically approximated during the calculation of the gradients of the model parameters. MCMC techniques are commonly used to perform this approximation by drawing samples from the model distribution \cite{bremaud_markov_2020,bishop_pattern_2006,koller_probabilistic_2009}.

Many studies attempted to use alternatives to classical sampling methods: methods capable of generating samples rapidly without compromising the quality of the sample. As mentioned in the Introduction, the D-Wave QA has been investigated as a sampler for RBM training \cite{adachi_application_2015,benedetti_estimation_2016,dixit_training_2021,kalis_hybrid_2023,korenkevych_benchmarking_2016,rocutto_quantum_2021}. In those efforts, samples produced by the QA were used instead of the Gibbs sampling when calculating the expectation values under the model distribution of the log-likelihood gradient with respect to the model parameters. One property that makes the QA more attractive is the speed of obtaining samples; 1,000 or 10,000 solutions can be generated from a single D-Wave call almost instantaneously. In contrast, the Gibbs sampling and MCMC methods in general require many burn-in (thermalization) steps before a reliable sample can be reached. Another benefit that has long been expected from the QA is the superior quality of its samples; one of the logical expectations is its potential to reduce the impact of getting stuck inside a limited number of LVs, thereby enabling broader exploration of the configuration space.

Two main approaches of using the D-Wave for sampling are (1) the direct use of the D-Wave solutions obtained at an instance-dependent effective temperature of the hardware, with estimation of this effective temperature to calculate the model expectation values \cite{adachi_application_2015,benedetti_estimation_2016}, and (2) the use of the D-Wave solutions as initial seeds to start the MCMC chain \cite{korenkevych_benchmarking_2016,kalis_hybrid_2023}. In this work, the second approach was investigated. 

The following procedure was used to generate the seeds. Solutions returned by the D-Wave QA were used to find distinct LVs in the RBM energy landscape.  The corresponding LMs were then used as starting seeds for Markov Chains during Gibbs sampling. In this work, different strategies for incorporating QA-derived seeds were investigated and compared with CD-\textit{1} training. 

\subsection{Markov Chain Monte Carlo Methods}\label{sec:2.2}
In this work, the QA-based sampling is compared to the classical MCMC Gibbs sampling. As was mentioned in Section \ref{sec:2.1}, many statistical quantities cannot be computed exactly. MCMC is a family of stochastic sampling algorithms that can be used to estimate these intractable statistical quantities. The building block of these methods is the Markov chain. 

A Markov chain is a ``memoryless'' sequence of random variables $X_0,X_1,X_2,\ldots,X_n$, where the state of $X_n$ is dependent only on $X_{(n-1)}$ and not on the rest of the variables. This memoryless property can be written mathematically as:
\begin{equation}
p_{ij}^{(t)} = \Pr\left( X^{(t+1)} = j \,\middle|\, X^{(t)} = i \right)
\end{equation}
where $p_{ij}^{(t)}$ is the transition probability from state $i$ to state $j$ at time $t$, with $t >= 0$. If the transition probability $p_{ij}^{(t)}$ has the same value for all $t$, the chain is called a homogeneous Markov chain. 

An important variant of MCMC is Gibbs sampling. The Gibbs chain differs in how the transition probabilities are constructed. Given a joint probability distribution for a set of random variables denoted by $P_i$, where the range of variables is a discrete (or continuous) set, Gibbs sampling updates each random variable from the conditional distribution given all other variables to generate samples from this probability distribution. In the context of the RBM, the update of one hidden unit (variable) is performed according to the conditional probability given all the visible units (variables), and vice versa. More importantly, since there is no connection between units in the same layer (i.e., among hidden units or visible units), units in the same layer can be updated independently and simultaneously, making the sampling task faster.

Rigorously estimating the second term of the log-likelihood gradients (Eqs.~(\ref{eq3}) -- (\ref{eq5})) using samples obtained from the Gibbs sampling requires running the Gibbs chain long enough to reach equilibrium. Instead, running the Gibbs chain for only \textit{k}-step starting from the TPs provides a good approximation to the expectation values over $p(\upsilon)$ shown in the second terms of Eqs.~(\ref{eq3}) -- (\ref{eq5}) \cite{hinton_training_2002,bengio_greedy_2006,hinton_fast_2006,bengio_justifying_2009,hinton_learning_2007}. This is the idea behind the CD-\textit{k}) algorithm. Since one starts from TPs $\upsilon^{(0)}$, in each \textit{k} steps, $h^{(1)}$ is sampled from $p(h| \upsilon^{(0)})$, then sampling $\upsilon^{(k)}$ from $p(\upsilon| h^{(k-1)})$. The expectations are estimated with this $\upsilon^{(k)}$. One may expect that running CD-\textit{k} for only one step (i.e., $k=1$) should not be sufficient to reach equilibrium or even to provide a good estimation of the expectations over the $p(\upsilon)$. However, it was shown that only one-step contrastive divergence (CD-\textit{1}) is often enough to train the RBM model \cite{hinton_training_2002}. Nevertheless, the quality of the samples produced in this way and the speed of training are believed to be among the main problems of the classical RBM training. 

In this work, a different approach to initializing the Markov chain in the CD-\textit{k} training was investigated. As was mentioned in the Introduction, a typical outcome of an RBM training is the formation of LVs with LMs in the vicinity of one or another TP. This is one of the reasons why starting the Markov chain from one of the TPs during CD-\textit{k} (i.e., using TPs as seeds) is an efficient classical sampling strategy. However, early in training, LVs with LMs that are close enough to TPs in the configuration space may not be formed yet. Hence, an alternative approach to selecting seeds for the Markov chain may be desirable, especially at those early stages of the training. Using the D-Wave to generate these seeds is a promising approach. Furthermore, based on the previously reported complementarity between the D-Wave-based and the classical samples \cite{elyazizi1_compare}, the use of a hybrid QA/classical seed may be promising. In this work, the possibility of starting the Markov chain from a mixture of TPs and D-Wave-found LVs to improve the quality of the sample and, thereby, the trainability of the RBM model has been investigated. 

\subsection{Adiabatic Quantum Annealing and its Use for Sampling}\label{sec:2.3}
D-Wave Inc. was the first company to build a commercial QC, specifically, a QA. The QA hardware is an implementation of an Ising spin glass model with the energy function in the form:
\begin{equation}
E(s) = \sum_{i=1}^{N-1} \sum_{j=i+1}^{N} J_{ij} s_i s_j - \sum_{j=1}^{N} h_j s_j
    \label{eq7}
\end{equation}

\noindent where  $J_{ij}$ is the strength of the coupling between qubits $i$ and $j$, $h_j$ is the local field of  $j^{th}$ qubit, and $s_i$ is the spin of qubit $i$.

The intended purpose of a QA is to solve optimization problems by finding the global minimum (i.e., GS) of Eq.~(\ref{eq7}). The speed of the QA and, potentially, its ability to find GSs stems from how its search process differs from that of classical methods. Unlike Gibbs sampling or simulated annealing, which rely solely on thermal fluctuations, QA leverages also quantum-mechanical tunneling, where the transitions from one LV to another during the process known as adiabatic evolution are achieved primarily by tunneling through the energy barriers rather than jumping over them. 

While the main purpose of a QA is to find the GS, its inherently probabilistic nature often leads it to find other (i.e., excited) states while searching for the GS. For this reason, adiabatic QA has been investigated as an alternative to the MCMC methods, and it shows promise for sampling tasks due to its high sampling speed and potentially higher sample quality (e.g., the ability to sample high probability states hiding inside LVs having a narrow basin of attraction). However, as the D-Wave hardware continues to mature, its ability to find GS improves, making it less likely to find other states \cite{pelofske_comparing_2023}. This is expected to compromise its effectiveness as a sampler. Results from the previous work by the authors' group \cite{elyazizi1_compare} showed that the number of LVs found by the QA is inversely proportional to the probability of finding the ground state $P_{\text{GS}}$, which means that the diversity of the sample may suffer when the QA gets better at doing its main job of finding the GS. 

Nonetheless, in our prior work, the latest version of the QA was still found to be promising in delivering samples of reasonable variety. Furthermore, when comparing QA-based and classical samples using the approach of Ref. \cite{elyazizi1_compare}, the two sampling methods showed both some overlap and some complementarity. While many of the LVs were found by both methods, each sampling method missed some LVs found by the other.  The missed LVs were in mid- and high-energy (i.e., mid- and low-probability) regions, while the (undesirable) overlap between the two techniques was for low-energy (high-probability) LVs, i.e., the more important states. The usefulness of those sampling differences between the QA and MCMC remained an open question. Advancing understanding in this direction was one of the objectives of this work.

\subsection{Incremental Learning and Catastrophic Forgetting}\label{sec:2.4}
For neural networks (NNs) to be considered reliable, they need to be capable of learning continually; that is, the ability to learn different tasks over time. However, continual learning is challenging, since NNs are prone to CF. CF is a phenomenon from which NNs suffer when they are retrained on new data. Specifically, the adjustments made to the weights of the NN when learning new data gradually destroy what had been learned before. This challenge in ML is known as the stability-plasticity dilemma \cite{abraham_memory_2005}. While the process of learning new knowledge resembles human learning, humans do not forget past information as easily when new information is learned \cite{french_catastrophic_1999}.

In humans and mammals in general, CF is mitigated with the reactivation of past neural activation patterns (the process known as replay) \cite{mcclelland_why_1995,kumaran_what_2016,mcclelland_integration_2020}. Replay is a mechanism used in the human brain to retain previously learned knowledge via revisiting internal representations of past experiences.

Three main schemes have been proposed in the literature to alleviate CF in continual learning of NNs~\cite{parisi_continual_2019}. The first is regularization-based methods, which aim to contain the weights of the model during training~\cite{kirkpatrick_overcoming_2017,aljundi_memory_2018}. The second approach involves network expansion, which includes increasing the number of neurons in an existing layer or adding entirely new layers to the NN during retraining~\cite{rusu_progressive_2022,ostapenko_learning_2019,hou_lifelong_2018}. The third main scheme is replay-based learning, which involves storing  (or generating) representations of the previously learned data and mixing them with the new data during retraining~\cite{hinton_using_1987,robins_catastrophic_1993,robins_catastrophic_1995}.

There are two types of the replay: partial and generative. In the partial replay, a subset of previous training data is stored for later retraining. Many successful algorithms exist that rely on this method to alleviate CF~\cite{rebuffi_icarl_2017,hayes_replay_2021}. In the generative replay, instead of storing actual representations of the past data, a generative model, such as an auto-encoder or a generative adversarial network (GAN), is used to generate memories of previous training data, which are then mixed with the new data during training. Compared to the partial replay, which stores the previously learned data, the generative replay reduces memory requirements and alleviates privacy concerns associated with storing training data~\cite{hayes_replay_2021}.  Numerous algorithms implementing generative replay have been reported, applied to different types of generative models, including variational auto-encoders and GANs~\cite{rios_closed-loop_2019,he-bmvc2018,ostapenko_learning_2019,shin_continual_2017,wu_memory_2018,caselles-dupre_s-trigger_2021,draelos_neurogenesis_2017}.

Some of the key challenges in mitigating CF using generative replay include sample quality, scalability to more complex datasets, and mode collapse \cite{de_lange_continual_2022}. Mode collapse refers to the failure of generative models to produce sufficiently diverse samples from the data distribution, resulting in producing output that “collapses” to a limited set of samples. 

Given the D-Wave’s ability to efficiently generate a large variety of potentially relevant samples, it may be promising for efficiently generating samples belonging to specific desirable classes, which could be used in the mitigation of CF in IL. In this work, the QA was used for the generative replay-based mitigation of CF. This is the first known attempt to use QA in CF mitigation in continual learning. A protocol known as Task incremental learning (Task-IL) was adopted during the training of the RBM (See Section \ref{sec:3.3}).  

\section{Methods}\label{sec:3}
\subsection{RBM Embedding within the D‑Wave Lattice}\label{sec:3.1}
In this work, a larger RBM model was used than in the earlier investigations conducted by the authors’ group \cite{koshka_comparison_2020,koshka_toward_2020,koshka_comparison_2021}, but comparable to the authors’ latest contribution on investigating the statistical properties of the LVs sampled by the D-Wave~\cite{elyazizi1_compare}. This was possible by switching from the old Chimera hardware to the D-Wave’s Pegasus Quantum Processing Unit (QPU). Unlike Refs.~\cite{koshka_comparison_2020,koshka_toward_2020,koshka_comparison_2021}, where the embedding of the RBM model into the QA lattice was done and optimized manually, the current work leveraged D-Wave’s minor-embedding tools to accommodate the larger and more complex lattice of the Pegasus hardware. One of the challenges was to ensure that the ferromagnetic bonds’ strengths are significantly greater than the weights of the connections between the logical RBM units; this challenge remained at least as critical and difficult to ensure as in the previous work \cite{elyazizi1_compare}. In the current work, sampling from D-Wave was required at every training epoch, making it impractical (and infeasible) to search for an optimal scale factor (as it was done in Refs.~\cite{koshka_toward_2020,koshka_comparison_2021,elyazizi1_compare}) whenever the RBM distribution changed (i.e., every time when the weights and biases were updated during the training). Instead, a dynamic scale factor was used at every epoch ($ sf = \frac{\max\left( \left| \text{weights} \right| \right)}{\text{target} - \text{weight}} $). In addition to this external scaling, the D-Wave’s auto-scale was used.

To verify the quality of the embedding at different training epochs, a classification task was performed using the classically-trained RBM embedded into the D-Wave. As in Ref. \cite{elyazizi1_compare}, the classification error was found somewhat worse (i.e., higher) than in earlier results by the authors’ group, when the hardware with Chimera lattice was used  \cite{koshka_toward_2020,koshka_comparison_2021}: 8.3\% with Pegasus compared to 5\% with Chimera. In the current work, we relied on the optimal scale factor found in Ref.~\cite{elyazizi1_compare}. During the classification using the D-Wave, the visible units of the test patterns (handwritten digits) were clamped to the desirable values (i.e., the pixels values of the test patterns) by setting the bias of the corresponding qubit to $\pm4$ depending on the value of the pixel. The qubits corresponding to the labels were not clamped and thereby were allowed to find the optimal value during the quantum annealing, as was done during label reconstructions with classical RBMs. This clamping approach proved to be very reliable, with 99.7\% of test patterns maintained the clamped states of their qubits in the D-Wave solution, with only a single qubit flipping its state in the remaining patterns. Later, we found no effect of this rare qubit flipping on the classification error. We also examined the effect of enabling or disabling the D-Wave’s auto scale during the embedding and found that it had no significant impact. While the classification error with the Pegasus hardware was inferior to that in the earlier work conducted on the (smaller) Chimera lattices  \cite{koshka_toward_2020,koshka_comparison_2021}, the quality of the embedding was found satisfactory for the objectives of the current work, which is to establish if statistically significant improvements in training can be achieved.

\subsection{RBM Training }\label{sec:3.2}
\subsubsection{RBM Training by Contrastive Divergence} \label{sec:3.2.1}
As in Ref.~\cite{elyazizi1_compare}, the OptDigits handwritten digits dataset was used for RBM training in all the experiments in this work. The original dataset was scaled down from $28\times28$ pixels to $8\times8$ pixels. Ten additional pixels were added to represent the class labels using one-hot encoding, bringing the total number of required RBM visible units to 74. The patterns were then binarized for a Quadratic Unconstrained Binary Optimization (QUBO) representation (note that the Ising representation was used in Refs. \cite{koshka_comparison_2020,koshka_toward_2020,koshka_comparison_2021}). The training was performed using 1,000 training patterns and one \textit{k} contrastive divergence step. The use of multiple \textit{k} steps was also investigated. However, as should be expected, the differences between different sampling methods (which in this work were the QA-based and the classical Gibbs sampling) have been found to significantly decrease for longer MCMC chains. For that reason, only one \textit{k} step was used in most of the training experiments as a more promising setting to reveal any possible effect of the modest sampling differences on RBM training outcomes. 

To ensure that the quality of training was preserved after embedding an RBM model into the D-Wave hardware, an aggressive \textit{L2} regularization method was used during the update of weights and biases, similar to how it was done in Ref.~\cite{elyazizi1_compare}. The purpose of this was to constrain the weights and biases within the range of values supported by the QA solver. 

\subsubsection{D-Wave-assisted Sampling from the RBM}\label{sec:3.2.2}
During the D-Wave-assisted training, sampling from the RBM was conducted once at the beginning of each training epoch. A single mini-batch containing all the TPs was used. This approach allowed performing a single D-Wave call per training epoch to generate a sufficient number of samples for the RBM training. In contrast, splitting the data into numerous mini-batches would require multiple expensive D-Wave calls in each training epoch. 

The following approach was designed to implement the hybrid sampling. In selecting seeds for Markov Chains, a mix of classically-generated seeds and D-Wave-generated seeds was used. The classical seeds were similar to what is used in the classical CD-\textit{k}. However, instead of using all the TPs as classical seeds, we used a subset of randomly selected (following a uniform distribution) TPs from the training dataset to satisfy our choice of having only half of all the seeds (i.e., only half of the sampled states) be produced classically. For the seeds (and, therefore, sampled states) produced using the QA, the approach adopted in this work was the one the authors considered the most likely to amplify the possible (best case) advantages of the QA sampling, those hypothesized based on the results of Ref.    \cite{elyazizi1_compare}, as discussed in the Introduction. The thermalization was initiated from the bottom of different LVs found by the D-Wave. To maximize the sample variety, as many different LVs as possible were used. The LVs (and the corresponding LMs to serve as the seeds) were found from the D-Wave solutions obtained in a single D-Wave job, using the following procedure. After the RBM model, after the particular training epoch, was embedded into the QA hardware, as described in Section \ref{sec:3.1}, 1,000 D-Wave solutions were collected from the D-Wave by executing 1,000 quantum anneals in a single job. Starting from each of the distinct solutions found by the D-Wave (i.e., 1,000 or fewer when some of the states were found by the D-Wave more than once), a classical MCMC chain was run at $T = 0$ to achieve deterministic relaxation to the bottom of the LV inside of which the given D-Wave solution-state was residing.  Only distinct LMs found at the end of the search were kept, using a bitwise comparison to discard the duplicates. The default annealing time (20~$\mu$s) and the auto scaling (ON) were used. To handle possible chain-breaks in each logical unit represented by multiple qubits, the majority vote was conducted after each quantum annealing run. 

When the number of LMs was less than the target number of seeds (i.e., the target number of sampled states, which was 500 in this work), some of the LMs were used more than once, selected randomly from all the found LMs with a probability following the Boltzmann distribution at $T=1$. Conversely, when the number of LMs was greater than the number of required seeds, an additional selection method was used. In that case, to ensure that the D-Wave’s part of the sample is as close as possible to the Boltzmann distribution (which is the property expected from an ideal RBM sample), while also maximizing sample variety, LMs were selected randomly without replacement from all the found LMs, again following the Boltzmann distribution at $T=1$.

Some of the RBM training experiments used seeds found using only the D-Wave. The LM search and the seed selection procedures were similar to those used in the hybrid seed experiments.   

The classification error and the log-likelihood were used as two performance metrics to compare the regular CD-\textit{1} training, the training with D-Wave-only seeds, and the training with hybrid seeds. The classification task was performed as follows. Each test pattern with randomized labels was used to initiate a Markov chain, with a sufficient number of burn-in (i.e., thermalization) steps, followed by steps that were used to calculate the reconstructed label via majority vote. In all three cases, the trained RBM was tested on a test dataset consisting of unseen patterns. 

While the classification error is influenced by many different factors (e.g., generalization performance of the training algorithm), the log-likelihood ($\mathcal{L}$) is a measure of how well the model explains the training data. A higher $\mathcal{L}$ indicates a higher likelihood of the data under the probability distribution of the trained model.  To compute $\mathcal{L}$, the intractable partition function \textit{Z} in Eq.~(\ref{eq1}) was estimated using annealed importance sampling (AIS)  \cite{salakhutdinov_learning_2009}.  
 
\subsection{Catastrophic Forgetting Mitigation during Incremental Learning by RBM}\label{sec:3.3}
In the third part of this work, which focused on the mitigation of CF, a Task-IL protocol \cite{ven_generative_2019} was investigated. In this protocol, the model learned a subset of the classes in the dataset during consecutive tasks. This learning setup simulated the real-world scenario of continual learning of ML models. In this work, the first task involved training the model on digits from only two of the ten classes in the training dataset. Following the first task, the CF protocol without mitigation proceeds as follows. During each of the consecutive learning-forgetting tasks (i.e., all but the first task, which involved only learning), the RBM was trained on two new classes. The initial RBM model for each consecutive learning-forgetting task was the model trained in the previous learning-forgetting task, except for the first task, where the model was randomly initialized. Specifically, from Task 1 through Task 5, the model was trained on classes of handwritten digits  0 and 1, 2 and 3, 4 and 5, 6 and 7, and 8 and 9, respectively. As a result, the classes learned in earlier task(s) but excluded from subsequent task(s) experienced gradual forgetting. 

During this unmitigated forgetting, the classification error deteriorated (increased) when the RBM was tested on data from the classes that had been learned in the previous tasks but were then excluded from the training set in the current task. This CF behavior without mitigation is shown by the black curves in Fig.~\ref{fig5}. For example, in Fig.~\ref{fig5}a, the classification error of curve (A) No mitigation, when the RBM was tested on classes 0 and 1 (i.e., those that were the subject of forgetting in Task 2), increased from 0\% at the end of Task 1 to approximately 25\% after the training of Task 2 was completed. This classification error further deteriorated to around 40\% by the end of Task 5 in Fig.~\ref{fig5}a. 

Next, the CF protocol was supplemented with a forgetting mitigation strategy. The TPs used in each task were mixed with additional patterns. Specifically, the TPs from two classes of the main training dataset were supplemented with data generated from the model trained in the previous tasks (i.e., by “memories” of the data learned in the previous tasks). As a baseline, only classical MCMC-generated memories were first employed. The performance of this classical CF mitigation was then used as a reference to evaluate the QA-based CF mitigation. Prior to each learning-forgetting task, which is now referred to as a ‘learning-forgetting-mitigating’ task, digits of desirable classes (in our case, the classes that will be the subject of forgetting in each subsequent learning-forgetting-mitigating task) were generated (i.e., “recalled”) from the model trained in the previous task using MCMC. These generated digits were added to the digits that had been generated and used in the earlier learning-forgetting-mitigation tasks. Specifically, Task 1 was trained on classes 0 and 1, Task 2 was trained on 2 and 3, plus patterns (memories) of 0 and 1 classes generated from the model after the training of Task 1 was completed, Task 3 was trained on 4 and 5, plus memories of 0 and 1, and 2 and 3 classes generated from the model after completing the training of Task 1 and Task 2, respectively, and so on. Hence, Task 5 (the last task) was trained on classes 8 and 9, supplemented with memories of 0 through 7, each recalled from the model trained after their respective tasks.

To generate memories from the RBM model using MCMC, the following procedure was applied after training a task and before retraining for the next task. Randomized vectors were constructed, each with the same size as the patterns in the dataset. To ensure that the memories belonging to the desirable classes (i.e., the classes subject to CF in the next task) were generated, the label pixels of the vectors were clamped to represent one of the two target classes. Each of the random vectors with clamped labels then served as the initial seed for the Gibbs sampling, which was run 1,000 times per class, with each run conducting 200 \textit{k}-steps at temperature $T = 1$. This process resulted in 1,000 (not necessarily distinct) memories from each class. 

A novel way of using QA was to replace the classical MCMC-generated memories with QA-generated memories for the CF mitigation described in the previous paragraph. For pattern generation from a trained RBM model using the QA, an aggressive \textit{L2} regularization, following the procedure described in Section~\ref{sec:3.1}, was used during the training of the RBM to keep the weights and biases as small as possible compared to the strengths of the ferromagnetic bonds, to ensure the best possible embedding quality. After training, a QUBO matrix was then constructed using the scaled-down weights and biases. As in other parts of this work (Section~\ref{sec:3.1}), the RBM embedding into the QA hardware was done using D-Wave’s minor embedding tools. The following procedure was applied to generate a digit from a desirable class using the QA. The QUBO elements corresponding to the class labels for the class or classes to be generated were clamped by setting them to the maximum possible values of the qubit’s bias field ($\pm4$). Then, a single D-Wave call was made to obtain 1,000 solutions. Similar to Section~\ref{sec:3.2.2}, the default annealing time (20~$\mu$s) was used, and the majority vote scheme was chosen to handle chain breaks. However, the auto scaling was turned off in this experiment to avoid the possibility of the values of small weights falling below the hardware sensitivity limit.

For both the D-Wave- and MCMC-generated memories, two selection methods were used to select patterns from the batch of 1,000 generated samples to serve as memories in the CF mitigation by the generative replay. In the first method, during the Task \textit{i}, the generated patterns (i.e., the visible units of the D-Wave solutions or the visible units of the states generated by MCMC) were classified under the RBM model trained so far. From all those states that were classified as belonging to the two desirable classes (i.e., the classes that will be the subject of CF in the next Task \textit{i}+1), the $K_{\text{mem}}$ lowest-RBM-energy states were used as memories, where $K_{\text{mem}}$ is the required number of memories. In the second method, at each training epoch, $K_{\text{mem}}$  patterns were selected randomly from the correctly classified patterns independent of their RBM energy.

The RBM classification error was used as a metric to compare the performance of the MCMC-based and the D-Wave-based CF mitigation. Specifically, the classification was done at the end of each task of the Task-IL protocol using test patterns belonging to the classes subject to CF in that task and all the previous tasks. It means that after Task \textit{i} was completed, the testing was done using test patterns belonging to the classes that were used as the TPs in Task \textit{i}-1 and all the earlier tasks. For instance, after the training of Task 5 on classes 8 and 9, plus memories of the classes from previous tasks, the model is tested, separately, on pairs of digits from 0 through 7 classes. This approach allowed assessing how well the CF-mitigated model retains knowledge from past learning when trained on new data (classes). 

\section{Results and Discussion}\label{sec:4}
\subsection{Comparison of CD-\textit{1}, DW-seeds, and Hybrid Trainings}\label{sec:4.1}
As was mentioned in the introduction, one of the broader aims of this work was to verify the hypothesis that the absence of substantial improvements in BM training when using QA for sampling from BM probability distributions, as observed in previous studies \cite{benedetti_estimation_2016,korenkevych_benchmarking_2016,rocutto_quantum_2021,dixit_training_2021}, could be explained by the fact that the noticeable differences between the classical and the QA-based sampling are primarily in the regions of the configuration space that have a modest importance for the overall sample (i.e., states having medium-to-low probability). 

To address this question, the RBM training outcomes were compared in this work when using three different sampling approaches: standard MCMC samples as traditionally used in CD-\textit{1} training, samples produced by thermalization from seeds generated from the D-Wave solutions, and samples produced by thermalization from hybrid seeds combining classical and D-Wave-generated seeds. In addition to the CD-\textit{1} experiments with only one \textit{k} contrastive divergence step, many \textit{k} steps were also investigated. However, as should be expected, the differences between the three sampling methods have been found to significantly decrease for longer MCMC chains. For that reason, only one \textit{k}-step was used in most of the training experiments as the most promising setting to reveal any possible effect of the modest sampling differences on RBM training outcomes.

\begin{figure}[H]
\centering
\includegraphics{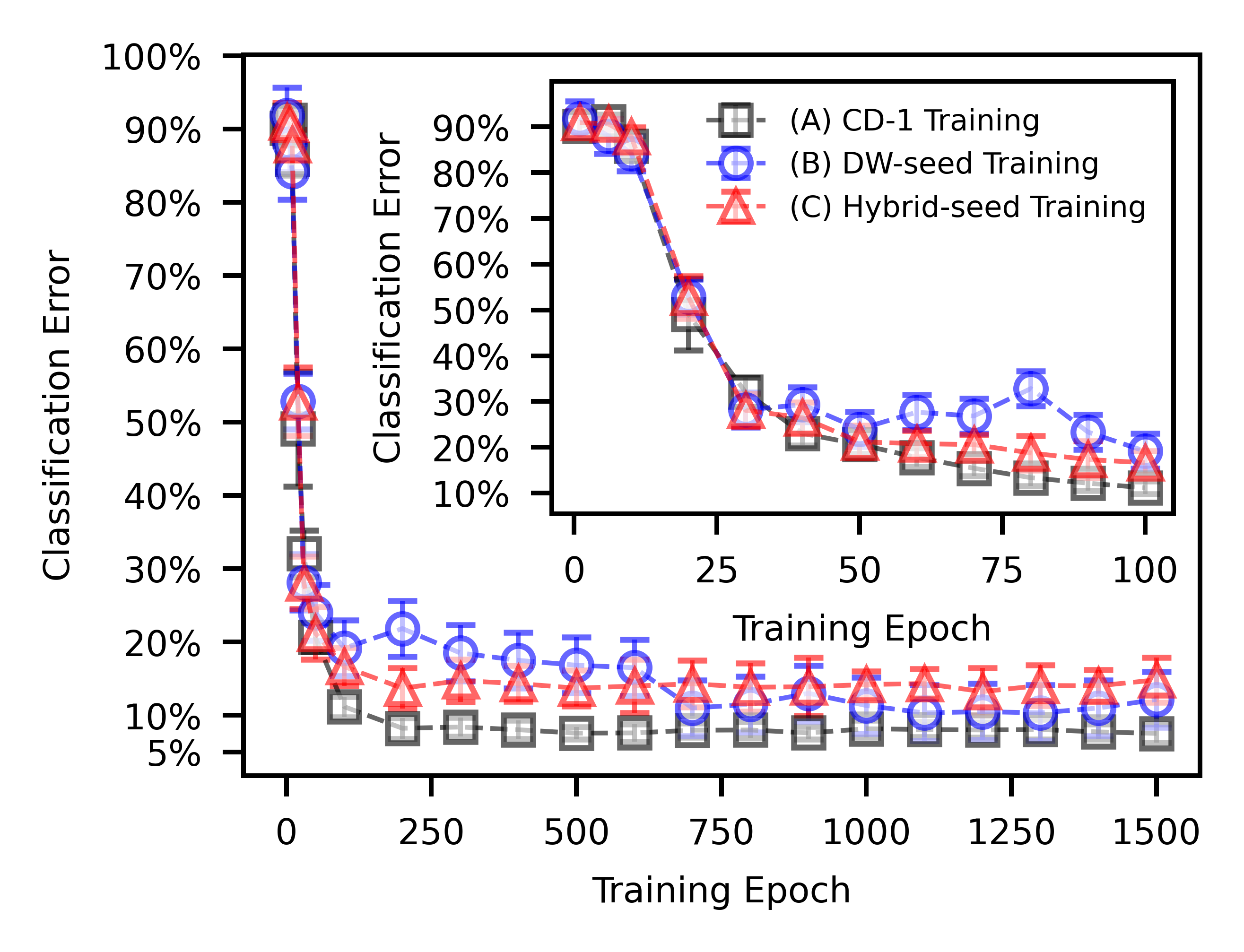}
\caption{The RBM classification error as a function of the training epoch, using three different sampling methods: (A) Gibbs sampling traditionally used in CD-\textit{1} training, (B) Markov chain initiated from D-Wave-generated samples as seeds, and (C) Markov chain initiated from hybrid seeds. Earlier in the training, the hybrid-seed sampling performed better than the DW-seed sampling. In the later stage of training, the hybrid approach slightly underperformed compared to the DW-seed method. The CD-\textit{1} training outperformed the other two sampling methods for most of the training epochs, although the DW-seed performed marginally better during the first 10 epochs.}\label{fig1}
\end{figure}

\begin{figure}[H]
\centering
%%%%%%%%%%%%Figure 2%%%%%%%%%%%%% width=0.9\textwidth
\includegraphics[]{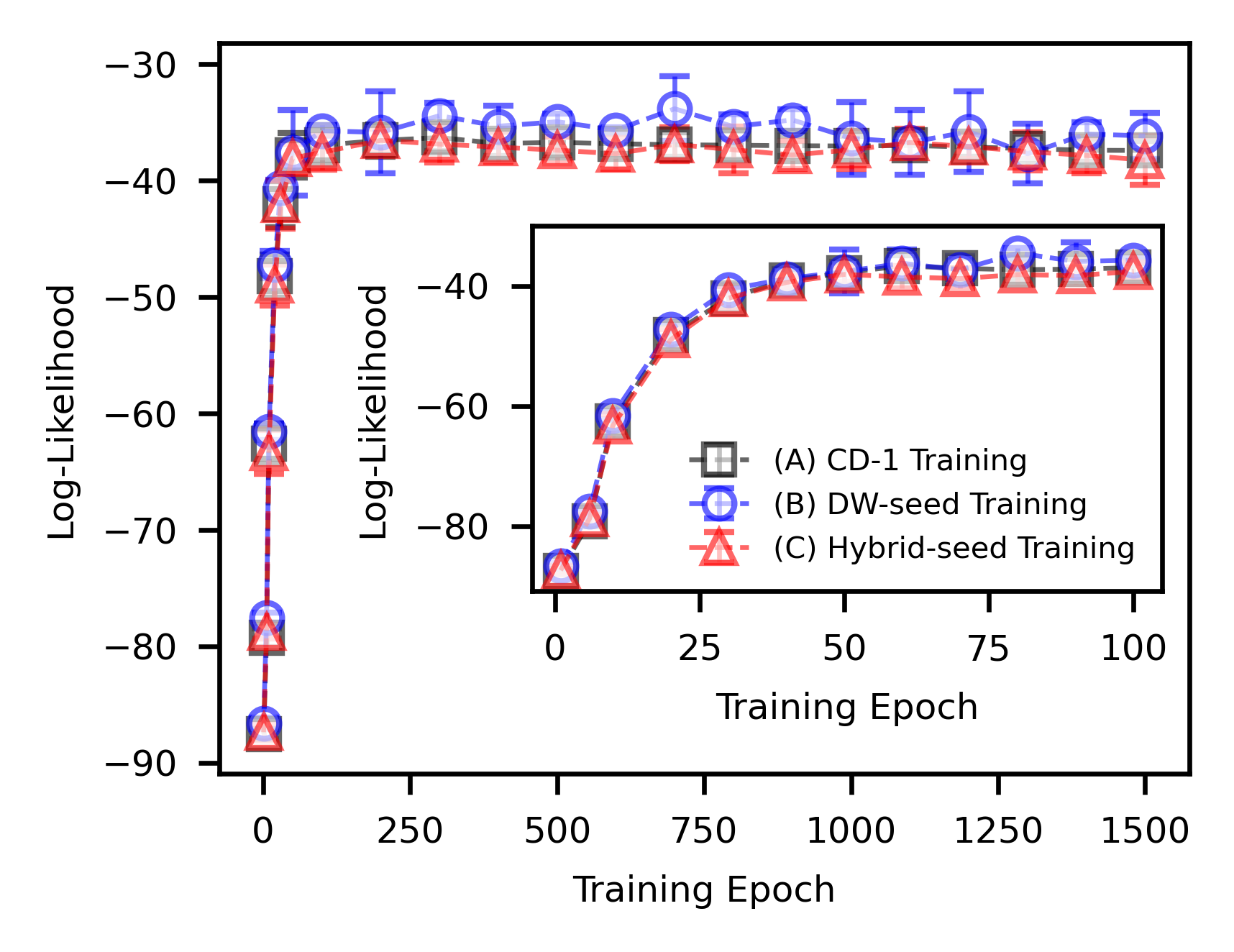}
\caption{The log-likelihood $\mathcal{L}$ of observing the training patterns under the probability distribution of the trained model,  shown as a function of the training epoch. The three curves are as in Fig.~\ref{fig1}: (A) classical Gibbs sampling, (B) Markov chains initiated from D-Wave solutions as seeds, and (C) Markov chains initiated from hybrid seeds. It follows from the figure that all three models result in close values of $\mathcal{L}$, with the DW-seed training showing a slight advantage compared to the other two over most of the training epochs.
}
\label{fig2}
\end{figure}

First, we discuss the results of comparing the three different sampling methods using the two metrics discussed in the Methods section: the classification error and the log-likelihood. Figures~\ref{fig1} and \ref{fig2} show these metrics as a function of the training epoch for the three different sampling methods, labeled as (A), (B), and (C): (A) the regular sampling of CD-\textit{1} training, (B) MCMC-based thermalization from the D-Wave-generated seeds (DW-seed training), and (C) thermalization from hybrid seeds. In (A), the Gibbs chain was initiated from each of the 1,000 TPs from the training dataset. In (B), LMs found by relaxation from D-Wave solutions, as described in Section \ref{sec:3.2.2}, were used as the start seeds of the Gibbs chain. The hybrid seeds in (C) were a combination of LMs found by relaxation from D-Wave solutions and from randomly selected TPs from the training Dataset (see Section \ref{sec:3.2.2}).  In (A) and (B), the results were averaged over 10 training runs. In (C), due to the limited D-Wave access time, the results were averaged only over 3 runs. 1,000 samples were used in (A), (B), and (C), which is equal to the size of the training dataset. In this work, in the hybrid training, LMs produced by the D-Wave constituted half of the total number of samples.  In all three cases, the trained RBMs were tested on the same test dataset consisting of unseen patterns. 
%%%%%%%%%%%%%%%%%%%%Figure 1
%%%%%%%%%%%%%%%%%%%%

As follows from Fig.~\ref{fig1}, early in the training, the hybrid training performed better than the DW-seed training. On the other hand, later in the training, the hybrid training performed slightly worse than the DW-seed training. The CD-\textit{1} training outperformed the other two approaches for most of the training duration, although the DW-seed performed slightly better during the first 10 epochs.

Figure~\ref{fig2} shows the log-likelihood, estimated by AIS, as a function of the training epoch for the same three sampling methods as in Fig.~\ref{fig1}.  As follows from Fig.~\ref{fig2}, all three methods resulted in close log-likelihood values, with the DW-seed training slightly outperforming the other two for the majority of the training duration. 

The result of the RBM training using the different sampling approaches in Fig.~\ref{fig1} and Fig.~\ref{fig2} revealed that, against the initial most optimistic expectations based on the authors’ prior work \cite{elyazizi1_compare}, the efficiency of the hybrid training failed to exceed that of the classical CD-\textit{1} training and was, in fact, somewhat worse. As expected, the potential for improvement is limited by the substantial overlap between the QA and the classical techniques and less complementarity than what would be required for producing significant improvements in the training. This is exactly because the differences between the sampling methods lie primarily in the lower probability region of the configuration space, which has only a modest impact on the quality of the samples \cite{elyazizi1_compare}. 

Furthermore, at later epochs of the training, the states corresponding to LMs are usually very close to one or multiple TPs. Hence, the conventional use of a TP as a seed to initiate the Markov chain of the Gibbs sampling should be expected to be sufficient at this later stage of training. The logical expectation could be that the D-Wave would still offer an advantage at earlier stages of training, when the traditional approach of selecting TPs as seeds is not yet a good starting point for a Markov chain. However, results from the previous work \cite{elyazizi1_compare} and this work shows that the D-Wave 
%%%%%%%%%%%%%%%%%%%%%%%%%%%%%
\noindent and MCMC overlap the most at this stage; both techniques find most of the same LVs, thereby limiting the practical benefit of QA-based sampling, even in the early stages of training when those benefits were expected to be the most likely.

\subsection{Additional Analysis of the Statistical Distribution of the Seeds with Respect to the LVs of the RBM Energy Function}

\begin{figure}[H]
\centering
     \includegraphics[]{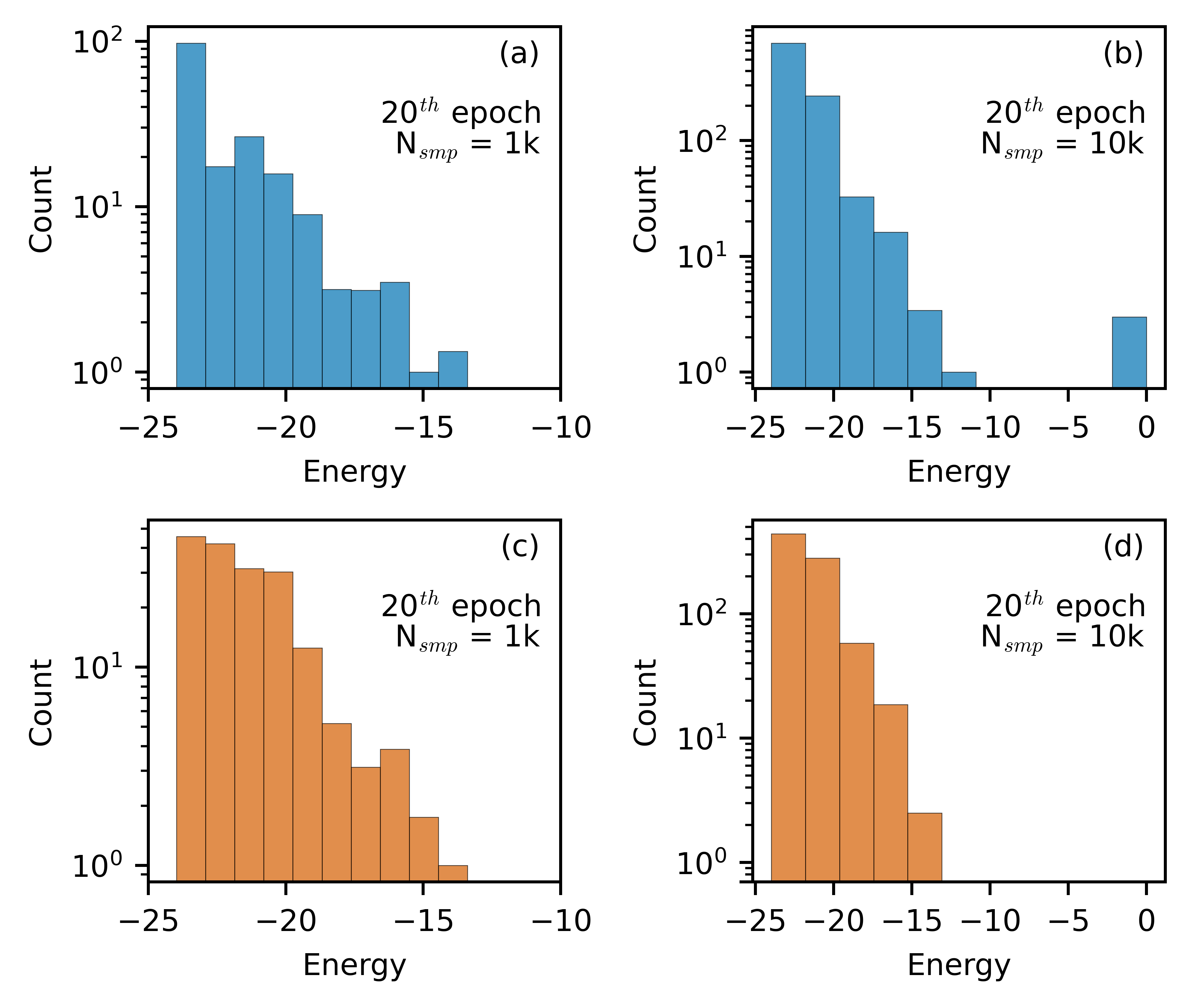} 
   \caption{(a) and (b): the average number of D-Wave solutions that belong to a single LV in a particular energy interval of the RBM energy of the LMs of those LVs to which those seeds belonged, shown for different energy intervals. (c) and (d): similar to (a) and (b), but for the average number of TPs that belong to a given LV in the particular energy interval. In (a) and (c), the number of samples $N_{\text{smp}}=$ 1,000, while in (b) and (d), $N_{\text{smp}}=$ 10,000 for both MCMC and the D-Wave.  All four plots are for the $20^{th}$ epoch. At this early stage of training, for both 1,000 and 10,000 sample sizes, more seeds end up in the same LV, compared to the later stage of training in Fig.~\ref{fig4}.
}
   \label{fig3}
\end{figure}

\begin{figure}[H]
  \centering
     \includegraphics[]{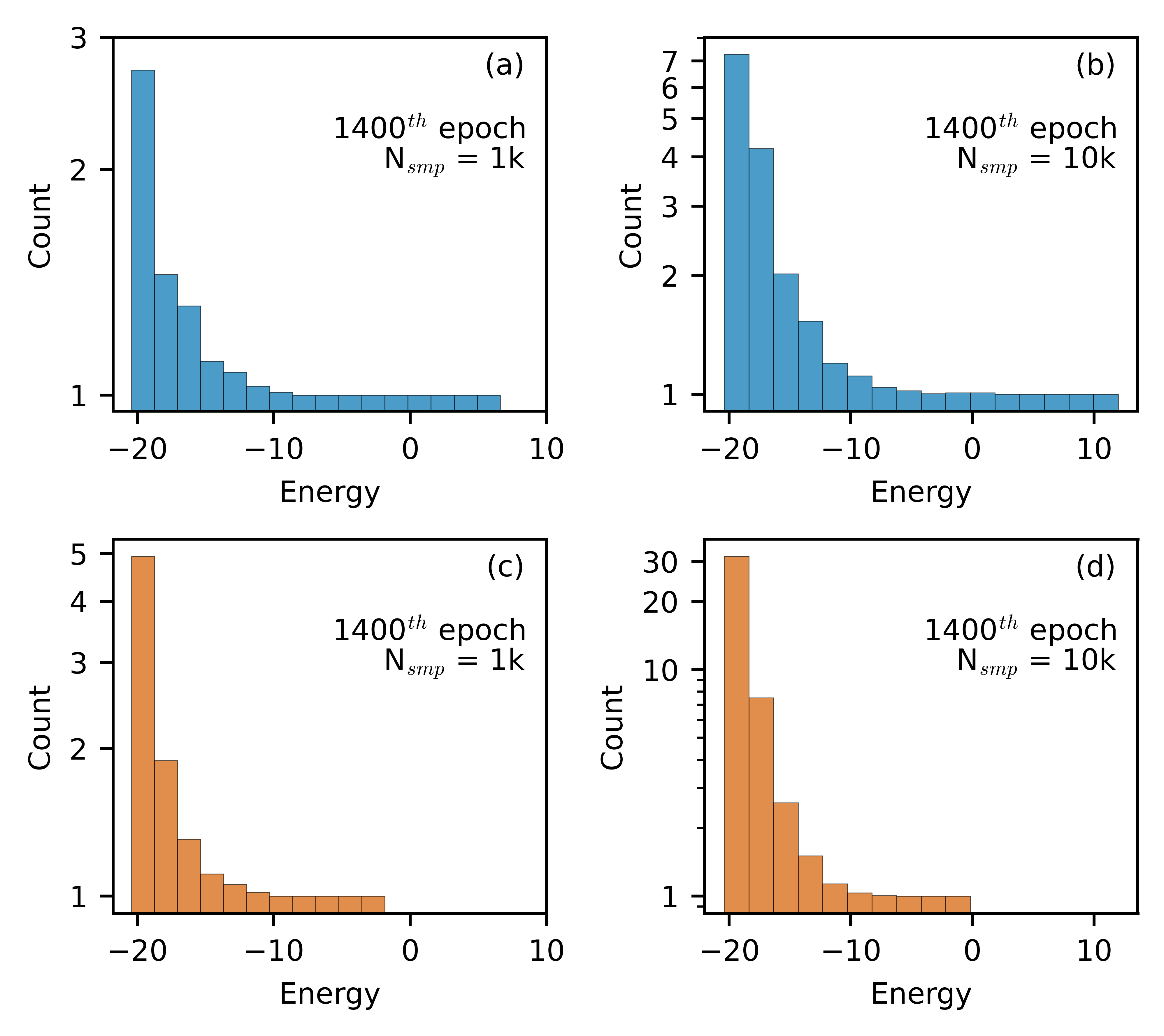} 
  \caption{Similar to Fig.~\ref{fig3} but for the 1,400$^{th}$ training epoch. The large number of samples in (b) and (d) resulted in more samples belonging to a single LV compared to (a) and (c). This higher number of samples (on average) belonging to a single LV is observed mostly in the low- and mid-energy ranges. For high energies, most LVs have one sample belonging to them for both sample sizes and sampling methods.
}
  \label{fig4}
\end{figure}

To gain further insight into the reasons why the hybrid sampling produced no improvement in RBM training, an additional analysis was conducted for the statistical distribution of the D-Wave solutions (those that were used to find LMs to serve as seeds) and their comparison to seeds used in the classical CD-\textit{k} training. Specifically, the average number of the initial D-Wave solutions that resulted in finding the same LV (i.e., D-Wave solutions belonging to a particular LV) in a given energy range was analyzed. This was compared to the average number of TPs belonging to a particular LV in the same energy range. The results are reported in Figs.~\ref{fig3} and \ref{fig4}.

The statistical trends observed in this analysis were found to be consistent with the prior observations of the LV statistics, even though at different sampling conditions \cite{elyazizi1_compare}. Lower-energy LVs contained more TPs and D-Wave solutions than higher-energy LVs; it could be as high as several hundred at the lowest energies early in the training, compared to higher-energy LVs, where it dropped to one or two (on average) TPs and D-Wave solutions per LV in an energy range. For LVs in the low-energy range, there were substantially more TPs and D-Wave solutions per LV early in the training than later in the training. The same trend was observed for mid-energy LVs. For high-energy LVs, there was only one or two (on average) D-Wave solutions or TP per LV for both early and late stages of the training. Further, early in the training ($20^{th}$ epoch, Fig.~\ref{fig3}), for both 1,000- and 10,000-sample sizes,  there were more D-Wave solutions than TPs per LV, which means that the D-Wave seeds explore a smaller number of LVs. For example, for the lowest energy range analyzed, on average, there were 700 D-Wave solutions and 425 TPs, for 10,000 samples (the lowest energy bars in Figs.~\ref{fig3}b and \ref{fig3}d). For the 1,000-sample case, there were on average 100 D-Wave solutions and 45 TPs per LV (the lowest energy bars in Figs.~\ref{fig3}a and \ref{fig3}c).

%%%%%%%%%%%%%% Figure 3%%%%%%%%%%%%%%%%%%%
%%%%%%%%%%%%%%%%%%%%%%%%%%%%%%%%%%%%%

Later in the training (1,400$^{th}$ epoch, Fig.~\ref{fig4}), for both sample sizes, in the lower-energy part of the spectrum, there were fewer D-Wave solutions per single LV in the D-Wave sampling compared to the number of TPs per single LV in MCMC. For example, in the lowest energy range analyzed, on average, there were 7 D-Wave solutions per LV compared to 30 TPs per LV for the 10,000-sample case (lowest energy bars in Figs.~\ref{fig4}b and \ref{fig4}d). Similarly, for the 1,000-sample case, there were 2.75 D-Wave solutions and 5 TPs per LV (lowest energy bars in Figs.~\ref{fig4}a and \ref{fig4}c). When analyzing the high-energy part of the spectrum, most of the found LVs had only one D-Wave solution belonging to them. The seeds used in the classical CD-\textit{k} training (i.e., the TPs used as the seeds) followed a similar trend (Fig.~\ref{fig4}). This is consistent with the prior observations that many LVs in this part of the energy spectrum remain not involved in the sample by either of the two techniques. 

%%%%%%%%%%%%%% Figure 4%%%%%%%%%%%%%%%%%%%
%%%%%%%%%%%%%%%%%%%%%%%%%%%%%%%%%%%%%
The standard CD-\textit{k} training algorithm predominantly samples states from the lower-energy part of the spectrum, which is the higher-probability region. This sampling behavior makes this part of the spectrum the most important. In this part of the energy spectrum, many TPs/D-Wave solutions used as seeds end up in the same LVs. As newer generations of D-Wave hardware improve their ability to sample the ground state more consistently \cite{pelofske_comparing_2023}, the number of unique low-energy LVs identified by the D-Wave (including those missed by MCMC) decreases further, reducing the diversity of the generated samples. The D-Wave seeds already significantly overlap with classical (MCMC) seeds in this important region. The substantive differences between the two techniques exist only in the high-energy/low-probability range, as was shown in Ref. \cite{elyazizi1_compare}, which is less significant for sampling. Therefore, combining the D-Wave found LMs with the classical seeds should not improve the sample quality. In fact, it may degrade sample diversity and quality by combining similar seeds, rather than utilizing the TPs as starting seeds to the Markov chain. These observations help explain the negative result of Section~\ref{sec:4.1} in attempting to enhance trainability through a hybrid sampling approach. 

\subsection{Catastrophic Forgetting Mitigation during RBM Training }
The third goal of this work, as stated in the Introduction, was to investigate the feasibility of using the D-Wave QA in the generative-replay method of mitigating CF during the continual learning of the RBM. Mitigation of CF by MCMC-based generative replay was used as a baseline in this investigation. 
The degree of forgetting and the mitigation performance were quantified using the classification error. 

\begin{figure}[H]
\centering
\includegraphics{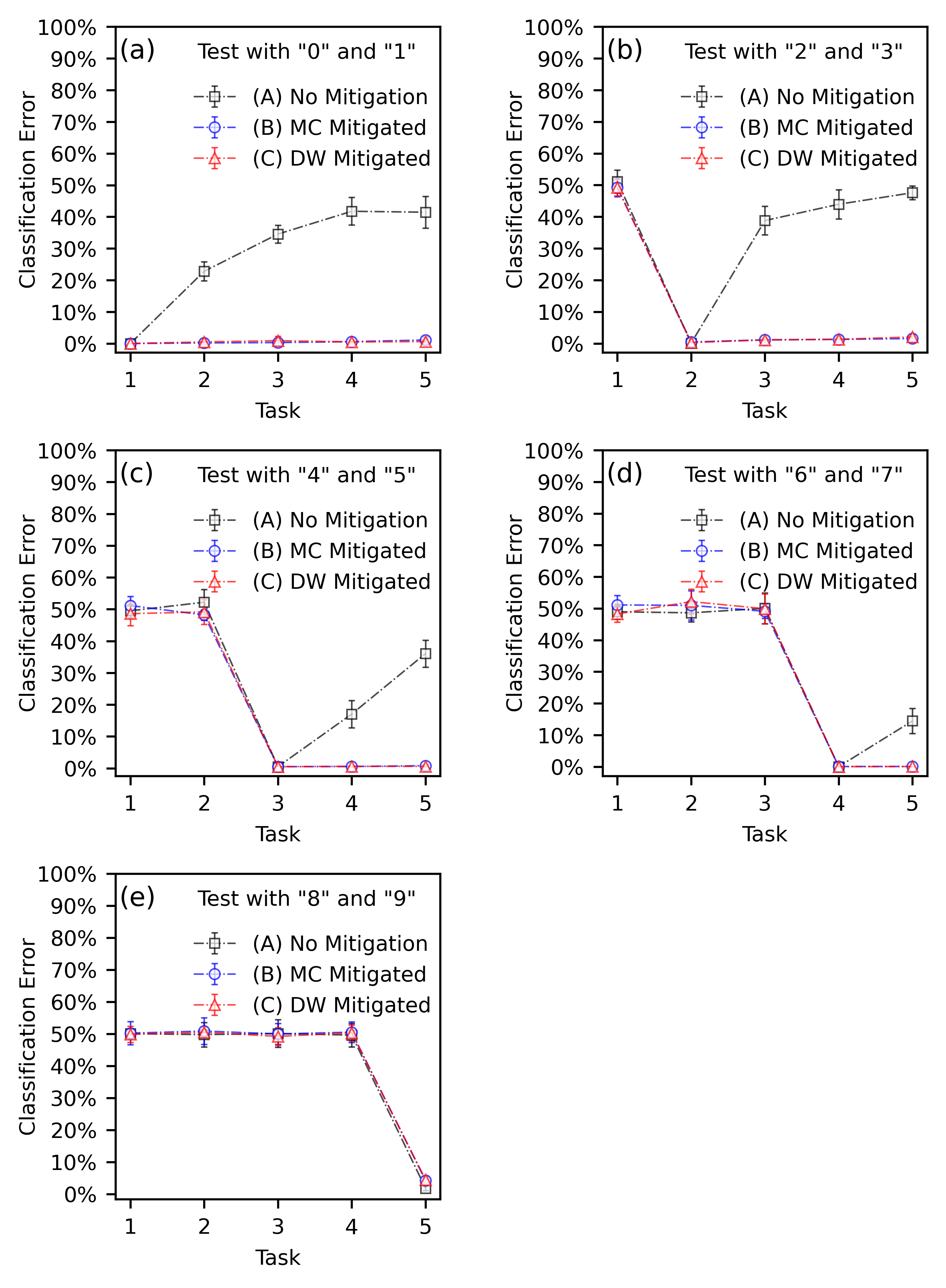}
 \caption{The classification error at the end of each task of the Task-IL RBM learning. The testing was done with patterns from two classes of handwritten digits that were used for training in one of the tasks: (a) classes 0 and 1 used for training in Task 1, (b) classes 2 and 3 used for training in Task 2, (c) classes 4 and 5 used for training in Task 3, (d) classes 6 and 7 used for training in Task 4, and (e) classes 8 and 9 used for training in Task 5. In each subfigure ((a) through (e)), the black curves (Curve A) are for unmitigated CF during training, the blue curves (Curve B) are for replay-based CF mitigation using MCMC-generated patterns (“memories”), and the red curves (Curve C) are for mitigation by a replay-based method using D-Wave generated “memories”. The patterns (the required number) from the 1,000 solutions generated from the model trained by the end of Task \textit{i}-1, to be added to the TPs for Task i and the later tasks, were taken as the visible units of states having the lowest RBM energy. The efficiencies of mitigating CF by the D-Wave and MCMC are comparable. However, the D-Wave generated that large number of “memories” much faster than MCMC.
}
 \label{fig5}
\end{figure}

Figure~\ref{fig5} 
shows the results of unmitigated CF (black curves (A)), the generative replay-based CF mitigation using D-Wave (red curves (C)), and its comparison to a similar mitigation using MCMC (blue curves (B)).

From Task 1 to Task 5 of the continual learning in each figure (the horizontal axes of Fig.~\ref{fig5}), the RBM was sequentially trained with digits from two classes: 0 and 1, 2 and 3, 4 and 5, 6 and 7, and 8 and 9, respectively. Each datapoint corresponds to the classification error at the end of each of the five tasks, after training on new data (data from two additional classes). The test datasets in each figure (from (a) through (e)) were different, and corresponded to pairs of classes learned in one of the five tasks, which then became the subject of forgetting: (a) classes 0 and 1 used for training in Task 1, (b) classes 2 and 3 used for training in Task 2, (c) classes 4 and 5 used for training in Task 3, (d) classes 6 and 7 used for training in Task 4, and (e) classes 8 and 9 used for training in Task 5. Therefore, each figure shows how well the RBM model retains knowledge of data learned in only one of the five tasks, after being trained on new data. Naturally, when, in some of the initial tasks, the model is tested on a pair that has not been used for training yet, the classification error is equal to 50\% (e.g., the datapoint for Task 1 in Fig.~\ref{fig5}b, the datapoints for Task 1 and Task 2 in Fig.~\ref{fig5}c, the datapoints for Task 1, 2, and 3 in Fig.~\ref{fig5}d, and the datapoints for Task 1, 2, 3, and 4 in Fig.~\ref{fig5}e).

The following method was used to select patterns to be added to the TPs in Task \textit{i} and the later tasks for generative replay mitigation. Those patterns were selected from the 1,000 solutions generated (using the D-Wave or MCMC) by the model trained at the end of Task \textit{i}-1. In Fig.~\ref{fig5}, the patterns (i.e., the required number) were taken as the visible units of states with the lowest RBM energy. Another selection method (not shown here) was also investigated, where the required number of patterns were drawn randomly from the correctly classified patterns among the 1,000 generated samples. 

The results of the CF without mitigation are discussed first. When the RBM is trained in a continual learning scenario without any CF mitigation, the classification error of Task \textit{i} (from the RBM model after training on Task \textit{i} data), when tested on
\noindent patterns from classes subject to CF (i.e., the data that had been used for RBM training in previous tasks), started to deteriorate (i.e., increased). As expected, the error of the unmitigated model (black curves (A)) keeps worsening as the model is trained on additional tasks, eventually reaching, in some cases, 50\% classification error, indicating that the classifier is merely guessing between the two classes in the test dataset. Specifically, for the black curves in Fig.~\ref{fig5}a, at the end of Task 1, the unmitigated classification error for classes 0 and 1 learned in that task is close to 0\%. When those classes are excluded from the training in the subsequent tasks, the unmitigated classification error increases to approximately 40\% by the end of Task 5.

In Fig.~\ref{fig5}b, showing the classification error for classes 2 and 3 learned in Task 2, the unmitigated classification error at the end of Task 1 (the black curve (A)) is 50\% (i.e., random), since RBM had not learned those classes in this task yet. At the end of Task 2, the unmitigated classification error for classes 2 and 3 learned in that task is close to 0\% (similar to that for classes 0 and 1 at the end of Task 1 in (a)). When those classes are excluded from the training in the subsequent tasks, the unmitigated classification error deteriorates to approximately 50\% by the end of Task 4. 

Similar unmitigated CF behavior is observed in Fig.~\ref{fig5}c through Fig.~\ref{fig5}e for classes 4 and 5 learned in Task 3, 6 and 7 learned in Task 4, and 8 and 9 learned in Task 5. Finally, in Fig.~\ref{fig5}e, there is no CF for the classes tested in this figure; the classification error simply shows the results of learning classes 8 and 9, with approximately 0\% classification error at the end of the experiment. 

The introduction of generative replay-based mitigations by means of generating patterns from the classes learned in Task \textit{i}-1 and earlier, and adding them to the training patterns of Task \textit{i} for training (as detailed in Section~\ref{sec:3.3}), is shown as blue and red curves in Fig.~\ref{fig5} for the MCMC- and the D-Wave-based mitigation, respectively. The mitigation significantly improved the classification accuracy (with almost 0\% classification error in some tasks) compared to the case without mitigation (the black curves in (a) through (e) in Fig.~\ref{fig5}). For example, in Fig.~\ref{fig5}a, by the end of Task $2$, the mitigation of forgetting classes 0 and 1 improved the classification error for those classes from 20\% to almost 0\% and remained at 0\% through Task 5. In Fig.~\ref{fig5}b, by the end of Task 3, the mitigation of forgetting classes 2 and 3 improved the error for those classes from 40\% to close to 0\% and remained at 0\% through Task 5. The results of the mitigation of other pairs of classes in (c) and (d) were similar. 

The performances of the CF mitigation by the D-Wave and MCMC were comparable to each other for most learning-forgetting-mitigating tasks, as shown by the near-overlap of the red and blue curves in Fig.~\ref{fig5}. Further, the two methods of selecting “memories” to be used for CF mitigation performed similarly, both for the D-Wave mitigated model and the MCMC mitigated model; the results of Fig.~\ref{fig5} and the results of the secondary pattern-selection method (not shown) were almost identical.  

In summary of this section, the use of QA in CF mitigation was implemented successfully and effectively eliminated forgetting. This result is encouraging, since this is the first known to the authors attempt to use the D-Wave QA for CF mitigation and one of the few instances of using the D-Wave as a generative model to produce samples from specific classes of interest in generative ML. 
\section{Conclusion }\label{sec:5}
Hybrid sampling for RBM training has been investigated in this work as an approach to understand and, possibly, address the lack of significant improvements (or no improvements) in the RBM training observed in previous research. The approach failed to achieve any substantial improvement in the training quality compared to the regular CD-\textit{k} training. In fact, the classification performance of the hybrid training, as well as the training utilizing only seeds produced by the D-Wave, were somewhat worse than that of the standard CD-\textit{1} training using Gibbs sampling, while the results for the log-likelihood were comparable between the three sampling methods. The reasons for the inferiority of the approaches utilizing the QA in this work have not been unambiguously established and is tentatively attributed to the less-than-perfect embedding of RBM into the QA. This is a problem that could not be fully resolved in this work by the weight decay, the use of the weight scaling factor, auto scaling, and other methods that proved efficient in the earlier publications by the authors’ group, which, however, used the smaller-scale Chimera rather than the Pegasus D-Wave architecture \cite{koshka_comparison_2020,koshka_toward_2020,koshka_comparison_2021}. In general, this outcome aligns with the trends reported in many previous studies of RBM training utilizing D-Wave sampling.  As was established in Ref.  \cite{elyazizi1_compare}, the RBM and the classical samples significantly overlap during the early stages of training. For that reason, in the earlier training epochs, combining seeds from the two methods or substituting the classical sampling with D-Wave-based sampling should not be expected to substantially affect the sample quality and therefore the training outcomes. On the other hand, improvements at later stages of the training, which were expected based on the observed differences in the sampling statistics between Gibbs and D-Wave sampling, did not materialize, apparently because those differences between the two methods are mostly for higher energy/lower probability states. Despite the expectation that those states (and the differences in sampling them by the QA versus MCMC) may still contribute to improving the quality of the sample, this work was not successful in extracting any benefit (or any significant effect) from those states on the sample quality. 

The new architecture of the Pegasus hardware allowed us to investigate an RBM with a size similar to that used in our most recent work ($74\times74$)  \cite{elyazizi1_compare}, and larger than what was investigated in earlier works by the author’s group \cite{koshka_comparison_2020,koshka_toward_2020}. While this larger QPU allows for embedding more practical models, the quality of embedding in this work was inferior. This may explain the differences in the classification error between this work and Ref.  \cite{elyazizi1_compare} compared to Refs. \cite{koshka_toward_2020,koshka_comparison_2021}, when the D-Wave-based generation of labels was used for classification. The reduced quality of the embedding may be further degrading the quality of the samples.

For the catastrophic forgetting part of this work, to the best of the author’s knowledge, the results are the first successful instance of using QA as a generative model for mitigating CF during continual learning of RBMs and, actually, the first instance of using QA for CF mitigation in generative ML. The performance of the D-Wave-based mitigation of the CF was similar to that of the mitigation utilizing classical MCMC. However, it is important to note that the D-Wave generates patterns much faster than the classical method. This may prove to be an advantage for larger RBMs (as well as other ML architectures) for which generation using MCMC becomes computationally intensive. If the D-Wave’s hardware scales sufficiently to accommodate these larger model sizes, the time advantage (even without the performance advantage) may be decisive. This is particularly promising when considered in the context of real-time training (e.g., a continuous flow of internet traffic) where the ability of the QA to generate a large sample belonging to a particular class nearly instantaneously may be both crucial and unique. Of course, in this work, the focus was on a critical, but only one, aspect of the CF mitigation:  the possibility of an efficient generation of a suitable sample. Other issues of CF mitigation, such as detection of when and what classes of training data need to be generated during mitigation, may require solutions outside the domain of QA (i.e., they will affect the classical, not the QA, part of the general algorithm) and therefore were beyond the scope of this work. 

The results of this work justify continuation of the research efforts to take advantage of the expected trends when the QC hardware matures, particularly as the number of qubits and couplers increases. Given the sampling speed advantage of the D-Wave over the MCMC, even with modest improvement in the sample quality, or even with no improvements, QA could still yield significant benefits. While the hybrid sampling approach explored in this study did not lead to substantial gains, future innovations in D-Wave-based sampling may prove more effective. Ultimately, QAs hold promise as a viable replacement for classical sampling methods.

Moreover, the differences between classical techniques and QA in sampling or generating states from difficult-to-reach regions of the configuration space may yet prove useful beyond just computation speed.  Many more ML applications could potentially benefit from QA used as a generative model. One example, explored in this work, is the use of QA in the CF mitigation. 

\bmhead{Acknowledgements}
This material is based upon work supported by the National Science Foundation under Grant No. CCF-2211841. Any opinions, findings, conclusions, or recommendations expressed in this material are those of the author(s) and do not necessarily reflect the views of the National Science Foundation.

\bibliography{references}

\end{document}